\documentclass[runningheads]{llncs}
\usepackage[T1]{fontenc}
\usepackage{graphicx}
\usepackage{graphicx}
\usepackage{xcolor}
\usepackage{comment}
\usepackage[utf8]{inputenc}
\usepackage{multirow}
\usepackage{arydshln}
\usepackage{subfigure}
\usepackage{float}
\usepackage{array}
\usepackage{tabularx}
\usepackage{graphicx}
\usepackage{url}
\usepackage{adjustbox}
\usepackage{xspace}
\usepackage{xcolor}
\usepackage{comment}
\usepackage{caption} 
\usepackage{adjustbox} %smn
\usepackage{booktabs}
\usepackage{hyperref}
\PassOptionsToPackage{bookmarks=false}{hyperref}
\usepackage{nicematrix}
\restylefloat{table}
\usepackage{latexsym}
\usepackage{amsmath, amssymb}
\usepackage{type1cm}        % activate if the above 3 fonts are
\usepackage{comment}
\usepackage{makeidx}         % allows index generation
\usepackage{multicol}        % used for the two-column index
\usepackage[bottom]{footmisc}% places footnotes at page bottom
\usepackage{newtxtext}       % 
\usepackage[varvw]{newtxmath} 
\usepackage{type1cm}        % activate if the above 3 fonts are
\usepackage{makeidx}         % allows index generation
\usepackage{graphicx}        % standard LaTeX graphics tool
\usepackage{multicol}        % used for the two-column index
\usepackage[bottom]{footmisc}% places footnotes at page bottom
\usepackage{newtxtext}       % 
\usepackage[varvw]{newtxmath}       % selects Times Roman as basic 
\usepackage{color,array}
\usepackage{comment}
\usepackage{subcaption}
\usepackage{soul,xcolor} % for \sethlcolor{}
\usepackage{mdframed} %for med frame

\newcommand{\orcid}[1]{\href{https://orcid.org/#1}{\includegraphics[width=8pt]{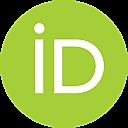}}}
\definecolor{highlightgreen}{rgb}{0.85, 1.0, 0.85} % Light green: correct
\definecolor{highlightorange}{rgb}{1.0, 0.9, 0.6} % Light orange: partially relevant
\definecolor{highlightred}{rgb}{1.0, 0.8, 0.8}  % Light red: hallucinated or irrelevant
\definecolor{highlightblue}{rgb}{0.85, 0.9, 1.0}  % Light blue: speculative or exaggerated
\definecolor{highlightyellow}{rgb}{1.0, 0.95, 0.5} % Slightly Deeper Yellow

\definecolor{backgG}{RGB}{255, 255, 153}
\definecolor{tagtxtG}{RGB}{102, 102, 0}
\definecolor{backgPc}{RGB}{179, 255, 179}
\definecolor{tagtxtPc}{RGB}{0, 102, 0}
\definecolor{backgPw}{RGB}{255, 179, 179}
\definecolor{backgPw}{rgb}{0.0, 1.0, 1.0}
\definecolor{tagtxtPw}{RGB}{0.0, 1.0, 1.0}
\definecolor{backgPo}{rgb}{0.76, 1, 1}
\definecolor{tagtxtPo}{RGB}{102, 0, 0}
\definecolor{backgPm}{rgb}{0.98, 0.81, 0.69}
\definecolor{tagtxtPm}{RGB}{0,1,1}

\makeindex 

\begin{document}
\title{Enhancing Business Analytics through Hybrid Summarization of Financial Reports}
\author{
Tohida Rehman \inst{1}\thanks{corresponding author} \orcid{0000-0002-3578-1316}
}

\authorrunning{Rehman et al.}
%\authorrunning{Bhattacharya, Rehman, Sanyal, and Chattopadhyay} 
\titlerunning{Rehman et al.}% Part of RIGHT running header
% If there are more than two authors, 'et al.' is used.%

\institute{
Jadavpur University, Kolkata, India.\\
\email{tohidarehman.it@jadavpuruniversity.in}
}

%\end{comment}

\maketitle
\begin{abstract}\unskip
Financial reports and earnings communications contain large volumes of structured and semi-structured information, making detailed manual analysis inefficient.  Earnings conference calls provide valuable evidence about a firm’s performance, outlook, and strategic priorities. The manual analysis of lengthy call transcripts requires substantial effort and is susceptible to interpretive bias and unintentional error. In this work, we present a hybrid summarization framework that combines extractive and abstractive techniques to produce concise and factually reliable Reuters\_style summaries from the ECTSum dataset. The proposed two-stage pipeline first applies the LexRank algorithm to identify salient sentences, which are subsequently summarized using fine-tuned variants of BART and PEGASUS designed for resource-constrained settings. In parallel, we fine-tune a Longformer Encoder–Decoder (LED) model to directly capture long-range contextual dependencies in financial documents.

Model performance is evaluated using standard automatic metrics, including ROUGE, METEOR, MoverScore, and BERTScore, along with domain-specific variants such as SciBERTScore and FinBERTScore. To assess factual accuracy, we further employ entity-level measures based on source-precision and F1-target. The results highlight complementary trade-offs between approaches: long-context models yield the strongest overall performance, while the hybrid framework achieves competitive results with improved factual consistency under computational constraints. These findings support the development of practical summarization systems for efficiently distilling lengthy financial texts into usable business insights.

\keywords{extractive summarization, abstractive summarization, natural language generation, financial documents, deep learning}

\end{abstract}

\section{Introduction}

The rapid growth of long and complex financial documents poses a significant challenges for efficiently extracting meaningful summaries. Transformer-based pre-trained models have demonstrated remarkable success across various NLP applications, including academic research, weather forecasting, social media analytics and many more. Motivated by these advances, this study applies pre-trained transformer models to domain-specific financial transcripts and reports to generate concise, bullet-point summaries that retain essential information.

Summarizing textual information is inherently challenging, and the exponential increase in data has made manual summarization impractical. In financial text summarization, the main task is to distill the most critical information from single or multiple documents into a brief, coherent and non-repetitive summaries.
Financial documents, in particular, are typically lengthy and often combine narrative text with tabular representations, further complicating the summarization process. Automatic text summarization aims to generate a condensed version of a document using computational methods that preserve its essential information. Generally, summarization techniques can be classified as extractive or abstractive. 
Extractive approaches identify and select the most important sentences or phrases from the source text, whereas abstractive approaches aim to understand the entire document to finds its main idea, similar to how humans does the summarization~\cite{el2021automatic}.

In the financial domain, summarization must not only shorten lengthy reports but also need to ensure factual accuracy. Hallucination or fabricated information poses a serious problem, as financial data is highly sensitive and error-prone interpretations can mislead analysts and investors.
In natural language generation, hallucinations are broadly divided into two categories. The first type, known as \textit{intrinsic}, occurs when the generated content contradicts or misrepresents information from the source. The second type, \textit{extrinsic}, appears when the model introduces additional details that cannot be verified against the original text~\cite{zhou2023comprehensive}. Such issues reduce the reliability and usefulness of automatically generated summaries, especially in domains like finance, science, medical and law, where factual precision is critical.

Large pre-trained language models achieve strong results in summarization, yet their computational demands pose challenges for long financial texts~\cite{zhou2023comprehensive}. Domain-specific fine-tuning provides a feasible alternative. This work addresses earnings call transcript summarization through hybrid extractive–abstractive approaches and a Longformer Encoder–Decoder (LED) model for handling long-context financial documents.

We evaluated our framework using the ECTSum dataset~\cite{mukherjee-etal-2022-ectsum}, which comprises long and unstructured earnings call transcripts paired with expert-written bullet-point summaries derived from Reuters articles. This dataset provides a realistic benchmark for assessing both content quality and factual consistency in financial summarization tasks.

%To improve factual consistency, we incorporate entity-level filtering during preprocessing~\cite{nan-etal-2021-entity}. 
The main contributions of this work are as follows:
\begin{enumerate}
\item We propose a hybrid summarization framework for resource-constrained settings that integrates LexRank-based extractive sentence selection with fine-tuned transformer models, including BART and PEGASUS, for summarizing long financial documents.
\item We also employed the Longformer Encoder-Decoder (LED)~\cite{beltagy2020longformer} pre-trained model to capture the complete essence of lengthy transcript summarization within the ECTSum~\cite{mukherjee-etal-2022-ectsum} financial dataset.
\item We evaluated all model performances using standard automatic metrics, including ROUGE \cite{lin2004rouge}, METEOR \cite{banerjee2005meteor}, MoverScore \cite{zhao-etal-2019-moverscore}, and BERTScore \cite{zhang2019bertscore}, along with its domain-specific variants, SciBERTScore \cite{beltagy2019scibert} and FinBERTScore \cite{araci2019finbert}. Furthermore, we assess factual consistency for each entity in the generated Reuters\_summary using precision-source, precision-target, recall-target, and F1-target metrics~\cite{nan-etal-2021-entity}.
\end{enumerate}

\section{Literature Review}

Text summarization has been extensively studied across a wide range of domains such as law, scientific research, biomedicine, news media, and social platforms, using both extractive and abstractive approaches~\cite{el2021automatic}.  In contrast, comparatively fewer studies have focused on financial text summarization, largely due to the length, complexity, and unstructured nature of financial documents.

Early work in this area focused on financial news summarization. De Oliveira et al.~\cite{de2002financial} developed a system based on lexical cohesion to generate concise summaries of Reuters financial news. Yang et al.~\cite{yang2003automatic} proposed a mobile-optimized system leveraging a fractal summarization model, which produced hierarchical and interactive summaries suitable for real-time financial information access in wireless environments. Filippova and Strube~\cite{filippova2009company} applied summarization to financial news from multiple sources for short-term trading analysis.

Subsequent studies began to address long-form financial documents using neural models. Zhang et al.~\cite{zhang2018extractive} proposed a hybrid extractive–abstractive framework (EAPC) that incorporated pointer and coverage mechanisms to reduce repetition and handle out-of-vocabulary terms. The integration of pre-trained word embeddings further improved semantic representation and summarization quality in both news and scientific domains~\cite{anh2019abstractive,rehman2023research,10172215}. Paul et al.~\cite{passali2021towards} introduced a financial news dataset consisting of approximately 2,000 Bloomberg articles paired with human-written summaries and demonstrated the use of deep learning models within an interactive summarization framework.

For financial report summarization, Agrawal et al.~\cite{agrawal2021hierarchical} proposed a multi-task model that jointly performs stock movement prediction and industry classification on 10-K filings. El-Haj and Ogden~\cite{el2022financial} applied a TF–IDF-based clustering method for extractive summarization using the FNS shared task dataset. La Quatra and Cagliero~\cite{la2020end} introduced a Sentence-BERT-based clustering approach to capture deeper semantic relations among sentences. Building upon these efforts, Shukla et al.~\cite{shukla2022dimsum} developed a T5-based extractive summarization model, later extending it with GPT-3 for the FNS shared task, which further improved summarization quality. Orzhenovskii et al. \cite{orzhenovskii2021t5} employed a pre-trained T5 model to generate summaries of financial documents, achieving strong results.

Vanetik et al.~\cite{vanetik2022summarization} proposed a neural extractive summarization approach that uses advanced sentence representations and document modeling to summarize long financial reports, outperforming baselines on the FNS 2021 dataset. More recently, Xie et al.~\cite{xie2024finben} introduced \textit{FinBen}, a large-scale benchmark comprising 42 datasets across 24 financial tasks for evaluating large language models.

Cardinaels et al.~\cite{cardinaels2018automatic} compared algorithmic and management prepared summaries of earnings releases, finding that automatic summaries were less positively biased, leading investors to make more conservative judgments. The ``ECT-BPS'' framework~\cite{mukherjee-etal-2022-ectsum} generated Reuters-style summaries by extracting and paraphrasing informative sentences from earnings calls. Tian et al.~\cite{tian2025template} proposed two LLM-based frameworks ``AgenticIR'' and ``DecomposedIR'' for summarizing templated financial reports, revealing a trade-off between brevity and comprehensiveness. ``ECT-SKIE''proposed  by Huang et al.~\cite{huang2025extracting}, employed a self-supervised, information-theoretic approach to extract key insights from earnings call transcripts. Chen et al.~\cite{chen2024term} proposed ``forward-looking claim planning'' for earnings calls, introducing methods to generate accurate, forward-looking arguments using transcripts collected from ``SeekingAlpha''.

In this work, we build upon prior research by evaluating hybrid extractive–abstractive summarization strategies and long-context transformer models on the ECTSum dataset under constrained computational settings, with a particular focus on factual reliability.

\section{Methodology}
In this section, we describe the pre-trained language models used in our fine-tuning pipeline. The following models were selected:

\subsection{Input Selection using LexRank}

To handle the token limits of pre-trained language models, we applied LexRank to extract the most salient sentences from earnings call transcripts. Each transcript was condensed into up to 15 key sentences, approximately 4,000 tokens. These sentences are used as input for fine-tuning the BART and PEGASUS models. This extractive step serves two purposes: it ensures that inputs remain within transformer token limits and concentrates the summarization process on the most salient content.

\subsection{BART}
\textbf{BART} \cite{lewis-etal-2020-bart} is a transformer-based sequence-to-sequence model that combines a bidirectional encoder, similar to BERT, with an auto-regressive decoder inspired by GPT. It employs activation functions such as GeLU instead of ReLU and incorporates several text corruption strategies, including token masking, text infilling, and sentence permutation, to improve its contextual understanding. BART used functions during pre-training as a denoising autoencoder. It helps to input documents are deliberately corrupted and the model learns to reconstruct the original text by minimizing cross-entropy loss between the predicted and reference sequences. This design enables strong performance across a range of natural language generation and other tasks. The model is available in two main configurations: \textit{BART-base} and \textit{BART-large}.
The \textit{BART-base} configuration consists of 6 encoder and 6 decoder layers, each with a hidden dimension of 768 and 12 attention heads. In contrast, the \textit{BART-large} variant expands to 12 encoder and 12 decoder layers, with a hidden size of 1024 and 16 attention heads.
The  \textit{BART-base} model contains approximately 139 million parameters. The \textit{BART-large} model contains approximately 406 million parameters. Both configurations demonstrate strong performance across various natural language generation tasks.

\subsection{PEGASUS}
\textbf{PEGASUS} \cite{10.5555/3524938.3525989} is a transformer based encoder–decoder model trained on massive text corpora using a pre-training objective. It is specifically designed for the task of abstractive summarization. It employs the Gap Sentence Generation (GSG) strategy, where multiple entire sentences mainly the most important ones are masked. Then generate as a single output sequence. This approach enables the model to capture summarization patterns more effectively than traditional span-masking objectives. 
The \textit{PEGASUS-base} architecture consists of 12 encoder and 12 decoder layers, each with a hidden dimension of 768 and 12 attention heads. 
In the \textit{PEGASUS-large} configuration, both the encoder and decoder contain 16 layers, with representations of hidden dimension 1024 and 16 attention heads. The \textit{PEGASUS-base} variant comprises around 223 million trainable parameters, whereas the \textit{PEGASUS-large} version includes roughly 568 million.
 
Both configurations are widely used for text summarization tasks.

\subsection{LED}
The Longformer Encoder–Decoder (LED) \cite{beltagy2020longformer} model is designed to efficiently process long input sequences. Its architecture combines sliding window and dilated sliding window attention mechanisms to extend the Transformer’s capacity for long-context modeling.
The model consists of both encoder and decoder stacks. Unlike standard Transformers that rely entirely on full self-attention, the encoder uses a combination of local and global attention structures. This design significantly reduces computational cost while preserving contextual understanding.
The decoder, in contrast, applies full self-attention to all encoded tokens and previously generated positions. Because pre-training LED from scratch is computationally demanding, \cite{beltagy2020longformer} initialized its parameters using those of BART, maintaining the same number of layers and hidden dimensions. This initialization allows LED to handle lengthy documents efficiently while inheriting BART’s robust linguistic representations. Fine-tuning further adapts the model through supervised learning, refining its ability to generate coherent and contextually accurate summaries while minimizing the need for extensive labeled data.

\begin{figure*}[!htbp] 
\centering
\includegraphics[width=0.96\linewidth]{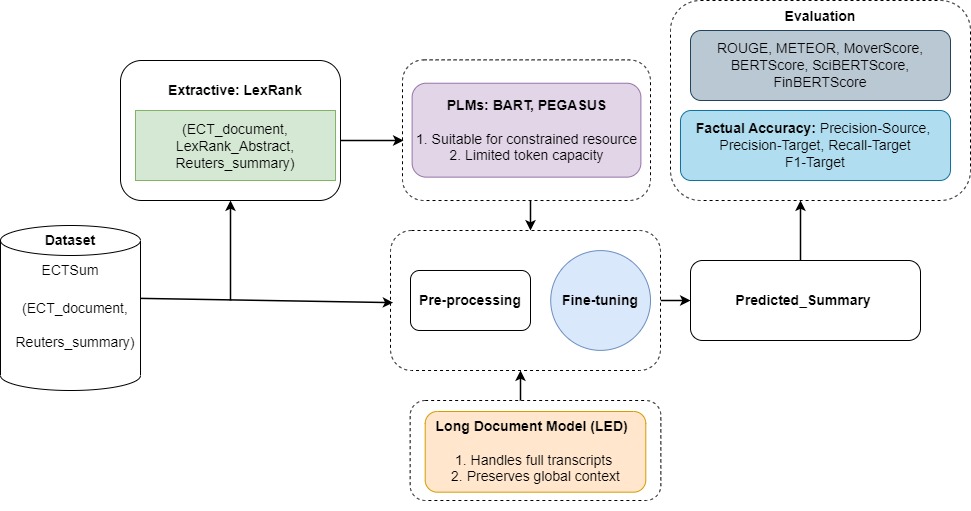}
\caption{Overview of the proposed hybrid extractive–transformer framework for financial document summarization.}
\label{fig:model_diagram}
\end{figure*}

Figure~\ref{fig:model_diagram} presents a schematic of the proposed hybrid summarization framework, where extractive sentence selection is followed by abstractive summarization, while a Longformer Encoder–Decoder (LED) is employed in parallel to handle long-context earnings call transcripts.

\section{Experimental Setup}
This section describes the dataset, implementation details including preprocessing steps, and evaluation metrics used in our study.

\subsection{Dataset}
We used the ECTSum dataset introduced by Mukherjee et al. \cite{mukherjee-etal-2022-ectsum}. The dataset contains Earnings Call Transcripts (\texttt{ECT\_document}) of publicly listed companies, collected from \textit{The Motley Fool}, a financial news platform that regularly publishes company earnings reports. The corpus includes transcripts from companies listed in the Russell 3000 Index, covering the period between January 2019 to April 2022.

Each record in the dataset consists of an \texttt{ECT\_document} paired with its corresponding \texttt{Reuters\_summary}, which is considered the ground-truth reference. The dataset was split into training, validation, and test subsets (70\%, 10\%, and 20\%), with 1,681, 249, and 495 samples, respectively.
%The dataset was divided into training, validation, and test sets in a 70:10:20 ratio, containing 1,681, 249, and 495 instances, respectively.

Based on simple whitespace-based segmentation, the average token counts for \texttt{ECT\_document} in the training, validation, and test sets were 2,860, 2,769, and 2,818. The corresponding \texttt{Reuters\_summary} entries contained an average of 44, 42, and 43 tokens, respectively.

\subsection{Implementation Details}

All experiments were performed on Google Colab using an NVIDIA T4 GPU. The setup was chosen due to limited computational resources.  

We first cleaned the dataset by removing extra whitespace and keeping only samples where the Earnings Call Transcripts (ECTs) had at least 20 tokens and the summaries had a minimum of 3 tokens. The generated summaries were restricted to less than 512 tokens, and input abstracts were allowed up to a maximum of 4096 tokens.  

Four pre-trained language models from the HuggingFace Hub were used: \texttt{BART-base}\footnote{\url{https://huggingface.co/facebook/bart-base}}, \texttt{BART-large}\footnote{\url{https://huggingface.co/facebook/bart-large}}, \texttt{PEGASUS-large}\footnote{\url{https://huggingface.co/google/pegasus-large}}, and \texttt{LED-base-16384}\footnote{\url{https://huggingface.co/allenai/led-base-16384}}. Each model was fine-tuned for three epochs.  

For \texttt{BART-base}, \texttt{BART-large}, and \texttt{PEGASUS-large}, training was done with  a learning rate of {\tt 4e-5}, batch size of 1, and a generation length of 128 tokens. The maximum input length was set to 1024 tokens with {\tt truncation=True} and output tokens to 128. For \texttt{LED-base-16384}, we used a batch size of 4 for both training and validation, with an input limit of 4096 tokens and {\tt truncation=True} and output tokens to 512.  

Before fine-tuning the BART and PEGASUS models, the LexRank algorithm was applied to extract the most relevant sentences from each ECT. For every transcript, LexRank selected up to 15 sentences with a total cap of 4,000 tokens. These extracted sentences were then used as input for fine-tuning.  

This step helped reduce document length while retaining the essential information needed for summarization. It also allowed efficient processing within a resource-constrained setup.

\subsection{Evaluation Metrics} \label{evaluation}

All fine-tuned model performances were evaluated using standard automatic summarization metrics: including ROUGE \cite{lin2004rouge}, METEOR \cite{banerjee2005meteor}, BERTScore \cite{zhang2019bertscore}, SciBERTScore \cite{beltagy2019scibert}, MoverScore \cite{zhao-etal-2019-moverscore} and FinBERTScore \cite{araci2019finbert}. These metrics assess the quality of the generated \texttt{Predicted\_summary} in comparison with the human-written \texttt{Reuters\_summary}. ROUGE measures word overlap. ROUGE-1 and ROUGE-2 capture unigram and bigram matches, while ROUGE-L identifies the longest common subsequence. METEOR evaluates sentence-level alignment by considering word order and synonyms. MoverScore and BERTScore assess semantic similarity using contextual embeddings. MoverScore uses Word Mover’s distance, whereas BERTScore calculates cosine similarity between BERT-based token embeddings. 

The modified metric SciBERTScore uses SciBERT, pre-trained on scientific texts, to capture more domain-specific information than general BERT. 
For financial summarization, we used FinBERTScore computed using ProsusAI/FinBERT available on Hugging Face\footnote{\url{https://huggingface.co/ProsusAI/finbert}}, a model pre-trained on financial texts to capture domain-specific semantics~\cite{araci2019finbert}.

While these metrics measure fluency and relevance, they may fail to capture factual consistency \cite{kryscinski-etal-2020-evaluating}. To address this, we used entity-level metrics from \cite{nan-etal-2021-entity} to verify factual correctness in the \texttt{Predicted\_summary}. Together, these metrics assess the lexical, semantic, and factual accuracy of summaries from \texttt{ECT\_document} to \texttt{Reuters\_summary}.

Let $\mathcal{N}(t)$ represent the named entities in the \texttt{Reuters\_summary}, $\mathcal{N}(h)$ the entities in the \texttt{Predicted\_summary}, and $\mathcal{N}(h \cap s)$ the entities shared between the \texttt{Predicted\_summary} and the \texttt{ECT\_document}. For multi-word entities, partial matches were accepted if any part of the entity appeared in the source. \textbf{Precision-source( prec\_s)}, calculated as,  
\begin{equation}
 prec_s = \frac{\mathcal{N}(h \cap s)}{\mathcal{N}(h)}
\end{equation}

This score represents how many entities from the \texttt{Predicted\_summary} are also present in the original \texttt{ECT\_document}. A lower $prec_s$ suggests hallucinated or missing entities. Entity-level accuracy with respect to the human-written \texttt{Reuters\_summary} is defined in terms of \textbf{Precision-target} (\texttt{prec\_t}), \textbf{Recall-target} (\texttt{recall\_t}), and \textbf{F1-score} (\texttt{F1\_t}) as follows:

\begin{equation}
prec_t = \frac{\mathcal{N}(h \cap t)}{\mathcal{N}(h)} 
\end{equation}

\begin{equation}
recall_t = \frac{\mathcal{N}(h \cap t)}{\mathcal{N}(t)}
\end{equation}

\begin{equation}
F1\_t = \frac{2 \cdot (prec_t \cdot recall_t)}{prec_t + recall_t}
\end{equation}

Here, $\mathcal{N}(h \cap t)$ denotes the common entities between the \texttt{Predicted\_summary} and the \texttt{Reuters\_summary}.  

These entity-based metrics can be computed in two forms. In the first, entities are treated as unique sets, ignoring repeated mentions. In the second, entities are treated as lists, where each occurrence is checked individually. Following \cite{nan-etal-2021-entity}, we denote the set-based version as $prec_s^{U}$, $prec_t^{U}$, $recall_t^{U}$, and $F1_t^{U}$, and the list-based version as $prec_s^{NU}$, $prec_t^{NU}$, $recall_t^{NU}$, and $F1_t^{NU}$.

\section{Results}

This section presents the experimental results on the ECTSum dataset. All fine-tuned models were evaluated using ROUGE-1, ROUGE-2, ROUGE-L, METEOR, MoverScore, BERTScore, SciBERTScore, and FinBERTScore. Table~\ref{Table:automatic_ECTSum} summarizes the performance of the models.

The \texttt{LED-base-16384} model achieved the highest overall performance, particularly for long transcripts, as reflected in ROUGE-1, ROUGE-2, METEOR, MoverScore and FinBERTScore. Both \texttt{BART-large} and \texttt{PEGASUS-large} performed competitively on the dataset with other metrics. \texttt{BART-large} achieved the best BERTScore and SciBERTScore, while \texttt{PEGASUS-large} obtained the highest ROUGE-L score.

When compared with the ECTSum benchmark reported by Mukherjee et al. \cite{mukherjee-etal-2022-ectsum}, which achieved ROUGE-L, ROUGE-1, and ROUGE-2 scores of 0.514, 0.467, and 0.307 respectively, our fine-tuned models obtained relatively lower scores.This difference can be attributed to limited computational resources, shorter input lengths, and fewer fine-tuning epochs. Despite these constraints, the LED model effectively captured the essential narrative flow of the financial transcripts and generated coherent, contextually relevant summaries.

\begin{table*}[!htpb]
\centering
\caption{\small Model performance of model-generated summaries ({\tt Predicted\_summary}) on the ECTSum dataset using standard metrics.}
\label{Table:automatic_ECTSum}
\begin{adjustbox}{width=1.0\linewidth}
{\begin{tabular}{|lcccccccc|} \hline
Model Name &ROUGE-1 &ROUGE-2 &ROUGE-L &METEOR  &MoverScore &BERTScore &SciBERTScore &FinBERTScore\\\hline
BART-base &30.40 &17.83 &27.15 &25.63 &20.76  &87.78 &69.01 &68.40\\ \hline
BART-large & 34.21 &19.25 &29.13 & 29.49 &23.48 &\bf{88.11} &\bf{69.62} &69.84\\ \hline
PEGASUS-large &34.40 &22.09 &\bf{31.80} &32.51 &22.29  &87.42 &69.32 &70.43\\ \hline
LED-base-16384 &\bf{37.29} &\bf{22.39} &31.10 &\bf{34.07} &\bf{24.27} &88.09 &69.42 &\bf{70.75}\\ \hline
\end{tabular} }
\end{adjustbox}
\end{table*}

Table~\ref{Table:entity-score-ETCSum} demonstrates the factual consistency evaluation of model-generated summaries on the ECTSum dataset. In all evaluations, the original \texttt{ECT\_document} was used as the reference input to ensure a consistent basis for comparison across models. 

The fine-tuned \texttt{LED-base-16384} model achieved the highest score in terms of precision-source ($prec_s^{NU}$, $prec_s^{U}$) and recall-target ($recall_t^{U}$) scores for both source and target entities. This indicates that it generated fewer hallucinated entities and maintained stronger alignment with the factual content of the input transcripts (\texttt{ECT\_document}).
However, its relatively lower F1-target scores suggest a trade-off between precision and recall, as the model tended to omit certain relevant entities while focusing on preserving core factual details. This behavior can be linked to its long-input handling mechanism, where attention is concentrated on prominent sections of the transcript, occasionally overlooking secondary information.

In contrast, the fine-tuned \texttt{PEGASUS-large} model obtained the highest precision-target ($prec_t^{NU}$, $prec_t^{U}$), recall-target ($recall_t^{NU}$), and F1-target ($F1_t^{NU}$, $F1_t^{U}$) scores. This performance reflects a better balance between factual coverage and entity accuracy, likely resulting from the use of ranking-based shorter input segments during fine-tuning, which enhances focus on key content while slightly reducing precision-source values.

\begin{table*}[!htpb]
\centering
\caption{\small Factual consistency evaluation of model-generated summaries ({\tt Predicted\_summary}) on the ECTSum dataset.}
\label{Table:entity-score-ETCSum}
\begin{adjustbox}{width=0.98\linewidth}
{\tiny
{\begin{tabular}{|ccccccccc|} \hline

Model Name &$prec_s^{NU}$ & $prec_s^{U}$& $prec_t^{NU}$& $recall_t^{NU}$ & $F1_t^{NU}$ & $prec_t^{U}$& $recall_t^{U}$& $F1_t^{U}$ \\ \hline
BART-base &74.02 &71.36 &60.85 &31.66 &35.08 &58.23 &34.65 &37.40\\ \hline
BART-large &75.47 &72.36 &58.73 &42.02 &40.97 &55.32 &42.79 &41.34\\ \hline
PEGASUS-large &72.35 &64.88 &\bf{67.75} &\bf{66.12} &\bf{56.05} &\bf{62.87} &42.25 &\bf{45.03}\\ \hline
LED-base-16384 &\bf{80.93} &\bf{77.49} &53.72 &49.12 &42.54 &50.55 &\bf{50.03} &42.71\\ \hline
\end{tabular} 
}
}
\end{adjustbox}
\end{table*}

\section{Case Study}
In this section, we present example cases illustrating the outputs of fine-tuned models on the ECTSum dataset. The analysis highlights both accurate and erroneous generations in the \texttt{Predicted\_summary} compared to the human-written \texttt{Reuters\_summary}. In each case, \colorbox{backgG}{yellow} marks factual errors or hallucinated entities. In this case study, the \colorbox{backgPo}{cyan} color highlights the correctly generated phrases or entity from the input that are not present in the \texttt{Reuters\_summary}.

%Figure \ref{fig:sample_Hallucination-CSPubSum1} shows an example of summaries generated by different models for the same \texttt{ECT\_document}.
Figure \ref{fig:sample_Hallucination-CSPubSum1} illustrates summaries generated by various models for the ECTSum dataset.
The fine-tuned \textit{BART-base} and \textit{BART-large} models produce a brief but factually correct summary, capturing part of the key information. The \textit{PEGASUS-large} model generates slightly more detailed summaries, successfully including additional accurate context from the transcript.  However, it also duplicates some phrases and occasionally repeats figures unnecessarily as ``q4 adjusted earnings per share \$2.94''.
\textit{LED} produces errors such as reporting ``\$5.70 per diluted share'' for Q4, which corresponds to full-year earnings per share rather than Q4, and is therefore factually inaccurate.
Additionally, the model incorrectly generated q4 adjusted pre-tax income of ``\$3.1 billion'’, whereas the actual fourth-quarter pre-tax income was \$668 million.

The case study demonstrates that summarization models can convey the overall financial context, but maintaining accuracy for numbers and named entities is still difficult. Mistakes, such as reporting incorrect earnings, underscore the importance of verifying figures and applying checks for entities and numerical consistency to ensure reliable summaries.
\begin{figure*}[!h] 
\centering 
\small
\begin{tabular}{ |p{12cm}|} \hline
%{\bf Input(abstract):} ``'' \\\hline filename :HRB_q3_2020
{\bf Reuters\_summary:} ``compname posts q4 earnings per share \$2.88.\\
q4 adjusted earnings per share \$2.94.''\\ \hline

{\bf BART-base:} ``q4 adjusted earnings per share \$2.94.''\\ \hline
  
{\bf BART-large:} ``q4 adjusted earnings per share \$2.94.\\ \hline

{\bf PEGASUS-large:} ``compname reports q4 adjusted earnings per share of \$2.94. \colorbox{backgG}{q4 adjusted earnings per share \$2.94.}''\\ \hline

{\bf LED:} ``compname reports q4 adjusted earnings per share of \$2.94.\\
\colorbox{backgPo}{compname posts fourth quarter 2021 adjusted earnings of \$1.3 billion}, or  \colorbox{backgG}{\$5.70 per diluted share}.\\
q4 adjusted pre-tax income of \colorbox{backgG}{\$3.1 billion}, \colorbox{backgPo}{an increase of \$26 million from the prior quarter}. \\
\colorbox{backgPo}{expects q1 global o\&p utilization rate to be in the mid-90s.}''\\ \hline
\end{tabular} 	
\caption{\small Comparison between the \texttt{Reuters\_summary} (ground-truth reference) and the model-generated \texttt{Predicted\_summary} from the ECTSum test dataset. The input \texttt{ECT\_document} and the \texttt{Reuters\_summary} are sourced from \texttt{\url{https://www.fool.com/earnings/call-transcripts/2022/01/28/phillips-66-psx-q4-2021-earnings-call-transcript/}}.}

\label{fig:sample_Hallucination-CSPubSum1} 
\end{figure*}
%2nd example

\section{Conclusion and Future Work}
Fine-tuned transformer models can generate coherent and contextually relevant summaries for long financial transcripts. The fine-tuned LED model achieved the highest precision-source and performed best on  METEOR, MoverScore, FinBERTScore, ROUGE-1, and ROUGE-2 scores. The fine-tuned BART-large model excelled in BERTScore and SciBERTScore, while PEGASUS obtained the best F1-target as well as ROUGE-L , balancing factual coverage and entity accuracy.

Future work can explore hybrid extractive-abstractive methods, enhanced attention for long inputs, and lightweight architectures to further improve accuracy and efficiency for real world  financial summarization tasks.

\bibliographystyle{unsrt}
\bibliography{llm_ref,finance_cite}
\end{document}